\documentclass[letterpaper]{article}
\usepackage{aaai2027} 
\usepackage[hyphens]{url}
\usepackage{array}
\usepackage{graphicx}
\usepackage{natbib}
\usepackage{amsmath}
\usepackage{amssymb}
\usepackage{booktabs}
\newcommand{\High}{\mathrm{High}}
\newcommand{\Low}{\mathrm{Low}}
\newcommand{\Unknown}{\mathrm{Unknown}}

\newcommand{\MemNone}{\ensuremath{\mathtt{Mem}_{\mathrm{none}}}}
\newcommand{\MemConv}{\ensuremath{\mathtt{Mem}_{\mathrm{conv}}}}
\newcommand{\MemSearch}{\ensuremath{\mathtt{Mem}_{\mathrm{search}}}}
\newcommand{\MemOKF}{\ensuremath{\mathtt{Mem}_{\mathrm{okf}}}}
\newcommand{\MemJev}{\ensuremath{\mathtt{Mem}_{\mathrm{jev}}}}
\newcommand{\MemProc}{\ensuremath{\mathtt{Mem}_{\mathrm{proc}}}}
\newcommand{\MemProcRL}{\ensuremath{\mathtt{Mem}^{\mathrm{rl}}_{\mathrm{proc}}}}
\newcommand{\MemJevProc}{\ensuremath{\mathtt{Mem}^{\mathrm{jev}}_{\mathrm{proc}}}}

\newcommand{\simnote}[1]{%
  \par\noindent\emph{#1}\par%
}
 
\begin{document}
\title{Procedure Inference as Memory: Learning Recruiter Decision Procedures
from OKF-Structured Episodes in Multi-Session Hiring Assistants}
\author{
    Praveen Kumar Bodigutla,
    Yizhi Zhang,
    Sai Vivek Kanaparthy,
    Arupe De   
}
\affiliations{
    \{pbodigutla, yizzhang, vkanaparthy, arde\}@linkedin.com
}
\maketitle
\begin{abstract}
Multi-session hiring assistants must infer evolving recruiter preferences and decision patterns across projects and time. We introduce Open Knowledge Format (OKF)-indexed episodic memory to organize interactions and retrieve related events through their lineage. We also introduce Procedure Memory, which learns reusable decision procedures from these episodes through prompted reflection or reinforcement learning. We compare these approaches with no memory, full-history context, and conversation search, and evaluate Jev-class models for index navigation and GPT-5.6 Luna as a lower-cost alternative to GPT-5.4.

In our evaluation, OKF-indexed retrieval achieves low-priority preference F1 scores of 0.93--0.95 at medium and high preference overlap between hiring projects. It significantly outperforms full-history context and conversation search in both settings, and no memory at high overlap, while using 69\% fewer context tokens than full history. Procedure Memory reduces conversational turns by up to 39\% relative to no memory. Qualitative and long-horizon evaluations provide evidence of temporal reasoning, with RL-learned procedures achieving 60\% on both correct abstention and exact-field identification. Jev-class models improve conversational efficiency but more frequently miss temporally relevant evidence, highlighting the importance of evidence selection.

On PersonaMem-v2, Procedure Memory with OKF retrieval achieves 50.2\% accuracy versus 48.6\% for full-history context and 29.0\% for no memory, using 90.5\% fewer reader input tokens than full history. GPT-5.6 Luna performs comparably to GPT-5.4 at lower cost. These findings highlight the value of structured episodic memory and learned decision procedures for efficient personalization and temporal reasoning.
\end{abstract}

\section{Introduction}

Memory-based personalization enables AI assistants to adapt to users' preferences and past interactions~\cite{salemi2024,jiang2025}. Recruiters interact with web-based hiring assistants~\cite{bodigutla2026,xu2026hierarchical} across multiple sessions and hiring projects, revealing explicit requirements and implicit priorities. For example, accepting a candidate who violates a stated requirement may indicate flexibility, while unsuccessful outreach may prompt recruiters to reconsider their requirements. These priorities evolve over time and may transfer across related projects. Effective personalization therefore requires assistants to infer not only recruiters' preferences, but also the decision patterns underlying their evolution.

Providing complete interaction histories at every model invocation is costly and introduces irrelevant or outdated evidence. Conversation search reduces context volume, but individual conversations often contain multiple activities, only some of which are relevant to a particular decision. Answering memory queries may require connecting dependent events and reasoning over their temporal relationships. For instance, identifying which preference a recruiter relaxed after an outreach rejection may require linking candidate attributes, earlier screening decisions, outreach outcomes, and subsequent recruiter actions across projects. These events may occur in different sessions, and temporal proximity alone does not establish causality. Prior work investigates implicit preferences, evolving memories, and context-sensitive retrieval~\cite{jiang2025,chen2026,zheng2026}. We investigate whether indexing individual activities and following their event lineage improves preference inference while reducing retrieval overhead.

For agents that accumulate memory across sessions, an additional challenge is distinguishing recorded observations from inferred preferences. Incorrect inferences may influence subsequent interactions and persist as the assistant continues adapting. We therefore investigate how structured episodic retrieval and learned decision procedures support reliable preference inference, particularly when evidence is incomplete or distributed across projects.

We introduce episodic memory structured according to the Open Knowledge Format (OKF)~\cite{okf}, representing recruiter activities as indexed, linked events. Rather than relying solely on similarity-based retrieval, our approach combines query-aware index navigation with structured event retrieval. An LLM interprets the query and selects relevant event identifiers from a compact OKF index. Lineage expansion recovers dependent activities, while BM25 and embedding-based retrieval rank supporting records for chronological interpretation. This enables the assistant to reconstruct relevant decision sequences at lower context cost while preserving event provenance.

Recent work on Jetriever~\cite{li2026jetriever} explores utility-based decision models for recovering indirectly relevant memories that similarity-based retrieval may miss. Since generative LLM-based index navigation introduces additional inference overhead, we investigate Jev-class models as an alternative for scoring and selecting relevant index entries. We use H2O-Lightning-4B~\cite{h2o2026lightning}, an open-weight, 4B-parameter decision model that produces probability-based selections in a single forward pass. Its compact architecture and low inference latency, with a reported median of 29 ms per decision on JevBench~\cite{mckinney2026jevbench}, make it a promising candidate for efficient index navigation. We retain the OKF index, lineage expansion, and downstream retrieval to examine whether faster selection preserves the evidence required for preference inference and temporal reasoning.

Building on episodic retrieval, we introduce \textit{Procedure Memory}: natural-language procedures learned from observable interactions to interpret recruiter decisions and infer evolving preference priorities. Unlike procedural memory for task execution~\cite{fang2025memp}, these procedures characterize recruiters' decision patterns. We investigate prompted reflection, which iteratively refines procedures without updating model parameters, and reinforcement learning (RL), which trains Qwen3.8-27B as an open-weight procedure learner using agreement with observed recruiter decisions as feedback.

Using simulated interactions, we compare no memory, full-history context, conversation search, OKF-indexed retrieval, OKF retrieval selected by Jev-class models, and OKF retrieval augmented with prompted or RL-learned procedures. We evaluate preference inference using priority F1, conversational efficiency using turns required to satisfy recruiter requirements, and context efficiency using retrieved tokens. A separate Day--15 long-horizon task tests whether the assistant can connect recruiter and candidate actions across projects to identify which preference was relaxed following outreach rejection, or correctly abstain when evidence is insufficient. This requires retrieving temporally related evidence and distinguishing genuine preference changes from unrelated actions. We further evaluate generalization on PersonaMem-v2~\cite{jiang2025} and compare GPT-5.6 Luna with GPT-5.4~\cite{openai2026pricing} to assess lower-cost inference.

In our evaluation, we find that OKF-indexed retrieval improves low-priority preference inference over full-history context and conversation search, achieving F1 scores of 0.93--0.95 under medium and high project overlap while reducing retrieved context tokens by approximately 69\% relative to full history. Procedure Memory further reduces conversational turns by up to 39\% relative to no memory and 28\% relative to OKF retrieval. In the long-horizon task, RL-based Procedure Memory achieves 60\% correct abstention and 60\% exact-field identification. Jev-class model navigation reduces conversational turns relative to LLM-based navigation but compromises flexible preference inference; without procedures, it achieves 0\% long-horizon field identification despite 90\% correct abstention. These results highlight the importance of retrieving temporally connected evidence rather than optimizing selection efficiency alone.

Evaluation on the external PersonaMem-v2 benchmark provides further evidence for the generalization of indexed retrieval augmented with Procedure Memory, achieving 50.2\% accuracy versus 48.6\% for full-history context and 29.0\% for no memory, while using 90.5\% fewer reader input tokens than full history. GPT-5.6 Luna performs comparably to GPT-5.4 across most evaluated settings, offering a lower-cost alternative~\cite{openai2026pricing}.

The remainder of the paper presents the simulation, memory strategies, OKF architecture, Jev-class model-based selection, and procedure-learning methods, followed by experimental setup, evaluations, results, conclusions and future directions.

\begin{figure*}[t]
\centering
\includegraphics[width=\textwidth]{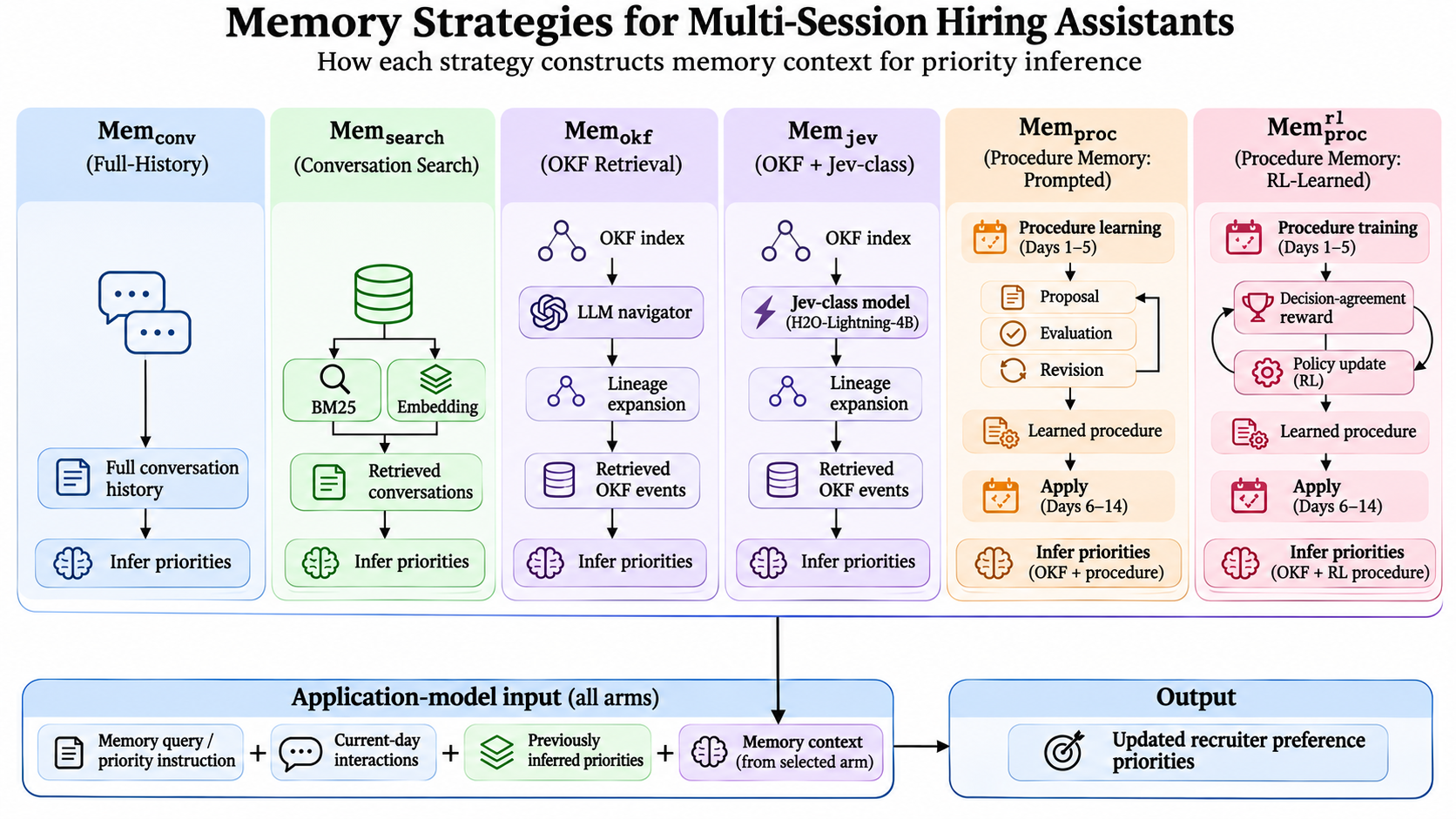}
\caption{Memory-context construction across strategies. Conversation and OKF indexes are updated incrementally; prompted and RL procedures are learned from observable Days~1--5 interactions. Jev-class models replace the LLM-based OKF event selector while retaining lineage expansion and downstream retrieval.}
\label{fig:memory-strategy-pipelines}
\end{figure*}
\section{Simulated Hiring Interactions}
\label{sec:simulated-interactions}

We simulate multi-session interactions between recruiters and a hiring assistant to study implicit preference priorities and their evolution following candidate outreach outcomes. The simulation covers job creation, preference specification, candidate review, outreach, and requirement revisions. Deterministic rules and templates govern workflow actions and state transitions, while Large Language Models (LLMs)\footnote{We use \emph{assistant} for the simulated hiring agent and \emph{LLM} for models performing memory navigation, interpretation, priority inference, and procedure learning.} support memory retrieval, priority inference, and procedure learning. Memory is scoped to each recruiter's projects and interactions, excluding other recruiters' histories. This controlled simulation, rather than a production-equivalent setup, enables evaluation against hidden recruiter preferences. Appendix~\ref{app:simulated-dialogue} illustrates the interaction flow.

\noindent\textbf{Recruiter Personas.}
Each recruiter has a persona governing requirement disclosure and interaction style. \emph{Fixed} personas provide a complete job description upfront, while \emph{exploratory} personas progressively specify or revise requirements. A hidden recruiter-specific relaxation attribute is shared across projects, enabling evaluation of cross-project preference inference.

\noindent\textbf{Latent Preference Priorities.}
Each recruiter--project pair is assigned values and latent priorities for five preferences: workplace type, location, skills, seniority, and visa sponsorship. The latent priority vector is $q_t\in\{\High,\Low\}^{5}$, while the assistant infers $\hat q_t\in\{\High,\Low,\Unknown\}^{5}$ from observable interactions. Preference values specify desired candidate attributes, while priorities determine whether mismatches are acceptable.

Recruiters archive candidates violating any $\High$-priority requirement and may accept or contact candidates satisfying those requirements. Revising a preference value sets its priority to $\High$. Unsuccessful outreach lowers the recruiter's designated preference from $\High$ to $\Low$, unless already $\Low$. This change is hidden from the assistant.

\noindent\textbf{Trajectories and Cross-Project Overlap.}
We simulate 30 recruiters, each with five hiring projects, across three initial overlap conditions: \textit{low} (median overlap well below $50\%$), \textit{medium} (approximately $50\%$), and \textit{high} (well above $50\%$), based on shared preference values and priorities. Each condition includes 10 recruiters with five trajectories each, yielding 150 trajectories. For each recruiter, trajectories share initial recruiter personas and preferences but vary candidate profiles, outreach outcomes, and preference revisions.

Each trajectory spans 14 daily sessions of up to 200 turns. Days~1--5 introduce one project per day, while Days~6--14 revisit existing projects. Conversation histories and memory indexes accumulate incrementally. At selected checkpoints, the assistant retrieves historical evidence to update inferred priorities, guiding subsequent candidate sourcing and interactions.

\section{Memory Strategies}
\label{sec:memory-strategies}

We compare five memory-based strategies against a no-memory baseline, yielding eight experimental arms when procedure-learning and Jev-class model-selection variants are included. All arms retain current-day interactions and previously inferred priorities. Figure~\ref{fig:memory-strategy-pipelines} illustrates their memory construction and retrieval pipelines.

\begin{itemize}
\item \textbf{No memory (\MemNone) --}
No historical interactions are retrieved.

\item \textbf{Full-history context (\MemConv) --}
All prior conversations are provided without filtering or ranking.

\item \textbf{Conversation search (\MemSearch) --}
Daily conversations are indexed as documents. BM25~\cite{robertson2009bm25} and embedding search each retrieve up to five documents\footnote{Retrieval limits selected by comparing different values of $k$.}, followed by deduplication.

\item \textbf{OKF-indexed retrieval (\MemOKF) --}
An LLM navigates the OKF index to identify relevant events. Lineage expansion recovers prerequisite activities, after which BM25 and embedding search each retrieve up to 25 eligible events\footnotemark[\value{footnote}]. Results are deduplicated and replayed chronologically.

\item \textbf{OKF retrieval selected by Jev-class models (\MemJev) --}
A Jev-class decision model replaces LLM-based index navigation by scoring indexed events against the memory query. Selected events anchor lineage expansion, followed by the same retrieval and chronological replay pipeline as \MemOKF.

\item \textbf{Procedure Memory (\MemProc{} and \MemProcRL) --}
OKF retrieval is augmented with natural-language procedures learned through prompted reflection or reinforcement learning (RL). These procedures guide interpretation of retrieved evidence and inference of evolving recruiter priorities. Current-project evidence takes precedence over conflicting cross-project patterns. A Jev-class model-selected variant (\MemJevProc) replaces LLM-based index navigation while retaining learned procedures and downstream retrieval.

\end{itemize}

\section{Episodic Memory with OKF}
\label{sec:episodic-memory}

We construct episodic memory using the Open Knowledge Format (OKF)~\cite{okf}, which organizes information through Markdown documents, YAML frontmatter, and links. Our implementation separates index navigation (\texttt{index.md}), event records (\texttt{events.md}), and dependencies (\texttt{lineage.md}). Each event records an observable recruiter or assistant activity, while \texttt{follows} links connect related activities, such as outreach responses and their preceding approvals. Events and relationships are appended incrementally, preserving historical records. Only observable activities, including recruiter feedback, outreach approvals, and candidate responses, are indexed.

\noindent\textbf{Context-aware event selection.}
Unlike conventional document retrieval, answering memory queries may require identifying events that are indirectly relevant through their dependencies and temporal context. Lexical and embedding similarity may miss such evidence, as explored by Jetriever~\cite{li2026jetriever}. We adopt a two-stage \emph{selection and retrieval} architecture. In \MemOKF{}, an LLM interprets the query and selects relevant event identifiers from the compact OKF index. In \MemJev{}, we use H2O-Lightning-4B~\cite{h2o2026lightning}, an open-weight Jev-class decision model, to score indexed entries for query relevance and select retrieval anchors. The Jev-class model scores each event for binary query relevance and selects the top 40 index entries as retrieval anchors before lineage expansion. Both approaches share the same OKF index and downstream retrieval pipeline. The exact decision prompt is provided in Appendix~\ref{app:jev-prompt}.

\noindent\textbf{Dependency-aware retrieval and provenance.}
Selected events are expanded through lineage links to recover prerequisite activities. BM25 and embedding-based search then retrieve eligible records, which are deduplicated and replayed chronologically. For example, interpreting an outreach rejection may require its preceding candidate profile and approval, alongside subsequent recruiter decisions retrieved as separate evidence. Event identifiers and lineage preserve provenance without asserting causal relationships. This design isolates event selection from evidence retrieval, enabling comparison of LLM- and Jev-class model navigation under the same retrieval pipeline. Appendix~\ref{app:okf-example} provides an annotated example.


\section{Procedure Memory}

OKF retrieval identifies \emph{which} episodes matter; \emph{Procedure Memory} addresses \emph{how} those episodes inform evolving recruiter priorities. An outreach rejection alone may not reveal which requirement was relaxed, while recurring cross-project patterns may suggest useful decision rules. We investigate prompted reflection, which iteratively refines procedures without updating model parameters, and reinforcement learning (RL), which trains an open-weight procedure learner. Both use agreement with observed recruiter decisions without accessing hidden priority labels; only RL updates model parameters. The Jev-class model-selected variant (\MemJevProc) changes event selection, not procedure learning. For example, a learned procedure may infer location flexibility when a recruiter initially archives candidates with location-only mismatches but later accepts such candidates, while retaining strictness for requirements supported by subsequent archival decisions.

\subsection{Shared Decision-Agreement Signal}

Given a procedure $\mathcal P$, target project $p$, and observable evidence $\mathcal H_{<t}$ at memory call $t$, a frozen application model infers priorities $\hat q_t=F(\mathcal P,\mathcal H_{<t},p)$. Evidence includes retrieved interactions and current-day context, excluding future events and hidden state. Procedures may use cross-project evidence but are evaluated against target-project decisions.

Let $W_t$ denote recruiter decisions before the next memory call. A deterministic predictor $g$ maps inferred priorities and candidate profile $x_d$ to \emph{archive} if any High-priority requirement is violated, and \emph{forward} otherwise (acceptance or outreach). Decision agreement is:

\begin{equation}
R(\mathcal P)=
\frac{\sum_t\sum_{d\in W_t}
\mathbf{1}[g(\hat q_t,x_d)=a_d]}
{\sum_t |W_t|},
\label{eq:procedure-reward}
\end{equation}

where $a_d$ is the recorded recruiter action. This score guides prompted reflection and serves as the RL reward.

\subsection{Prompted Learning through Reflection}

We use GPT-5.4 and GPT-5.6 Luna for procedure learning and priority inference. Learning comprises three proposal--evaluation rounds: an initial procedure from observable Days~1--5 interactions, followed by two revisions. Each procedure is applied at recorded memory calls and evaluated against subsequent recruiter decisions using Equation~\ref{eq:procedure-reward}.

Revisions receive previous procedures, scores, and feedback on low-scoring windows. We reserve 60\% of scoreable calls for refinement and 40\% for procedure selection, choosing the highest-scoring procedure and breaking ties in favor of the earliest attempt. The selected procedure remains fixed during Days~6--14.

\subsection{RL-Based Learning with an Open-Weight Model}

We train Qwen3.8-27B using Group Relative Policy Optimization (GRPO)~\cite{shao2024} to generate procedures from observable Days~1--5 interactions, with thinking disabled. For each prompt, GRPO samples $G=4$ procedures, each receiving decision-agreement reward $R_i$ from Equation~\ref{eq:procedure-reward}; malformed outputs receive $-1$.

We compute group-relative advantages without standard-deviation normalization:

\begin{equation}
A_i=R_i-\bar R,\qquad
\bar R=\frac{1}{G}\sum_{j=1}^{G}R_j.
\label{eq:grpo-advantage}
\end{equation}

For token $y_{i,s}$ and its preceding context $h_{i,s}$, the policy ratio is

\begin{equation}
r_{i,s}(\theta)=
\frac{\pi_\theta(y_{i,s}\mid h_{i,s})}
{\pi_{\mathrm{old}}(y_{i,s}\mid h_{i,s})}.
\end{equation}

The clipped policy objective with KL regularization is

\begin{equation}
\begin{aligned}
\ell_{i,s}(\theta)
&=\min\bigl(r_{i,s}(\theta)A_i,\\
&\quad \operatorname{clip}(r_{i,s}(\theta),
1-\epsilon,1+\epsilon)A_i\bigr),\\
\mathcal L(\theta)
&=-\mathbb E\left[
\left\langle\ell_{i,s}(\theta)\right\rangle_{i,s}
\right]
+\beta\mathcal L_{\mathrm{KL}}(\theta),
\end{aligned}
\label{eq:grpo-objective}
\end{equation}

where $\epsilon$ is the clipping threshold and $\langle\cdot\rangle_{i,s}$ averages over sampled completion tokens. The KL penalty measures divergence from a frozen reference learner:

\begin{equation}
\begin{aligned}
\mathcal L_{\mathrm{KL}}(\theta)
=\mathbb E\Bigl[
\Bigl\langle D_{\mathrm{KL}}\bigl(
&\pi_\theta(\cdot\mid h_{i,s})\\
&\|\,\pi_{\mathrm{ref}}(\cdot\mid h_{i,s})
\bigr)\Bigr\rangle_{i,s}
\Bigr].
\end{aligned}
\label{eq:grpo-kl}
\end{equation}

We estimate the KL term using a low-variance estimator and set $\beta=0.002$. Only the procedure learner is updated; GPT-5.4 remains frozen for priority inference, with reasoning effort set to \texttt{none}. Unlike prompted reflection, which uses recorded recall, RL reward evaluation uses trajectory-scoped lexical and embedding retrieval.

\section{Experimental Setup}
\label{sec:experimental-setup}
We conduct experiments in simulated multi-session recruiter interactions using Redis-backed lexical and embedding indexes and incrementally constructed OKF episodic memory. The compact OKF navigation index is approximately seven times smaller than its resolved event records. 
Days~1--5 provide observable interactions for procedure learning and selection, while learned procedures remain fixed during Days~6--14. Default priority-inference model is GPT-5.4, with reasoning effort set to \texttt{none}, to minimize inference latency in responsive user-facing applications. Experimental configurations are detailed in Appendix~\ref{app:experiment-config}.

\noindent\textbf{Preference-priority inference across memory arms.}
We compare five memory-based strategies against a no-memory baseline across low, medium, and high project overlap. Each memory arm operates within the same simulated hiring workflow, inferring recruiter priorities to guide candidate sourcing and subsequent interactions. We examine preference-inference accuracy, conversational turns, and memory overhead across Days~1--5 and Days~6--14. Procedure Memory uses procedures learned during Days~1--5 for the continuation period.

\noindent\textbf{Long-horizon behavior inference.}
We introduce a Day--15 long-horizon task requiring memory strategies to connect recruiter and candidate actions across hiring projects to identify which preference was relaxed following outreach rejection, or abstain when evidence is insufficient. We select the \emph{low-overlap} setting to make inference more challenging: limited similarity across projects requires identifying transferable behavioral patterns rather than simply copying preference priorities from other projects. The experiment uses a reworded evidence-seeking query and 20 arm-specific cases, balanced between answerable and insufficient-evidence cases. We examine correct abstention and exact-field identification separately. Experimental details and reference adjudication are provided in Appendix~\ref{app:abstention-protocol}.

\noindent\textbf{External validation on PersonaMem-v2.}
To investigate generalization beyond hiring, we use 255 held-out questions across 97 personas from PersonaMem-v2's \texttt{text/val} partition~\cite{jiang2025}. Three labeled questions per persona support prompted procedure learning and refinement over ten proposal and refinement rounds, scored by learning-question accuracy rather than recruiter decision agreement. Evaluation answers remain excluded from learning and selection. GPT-5.4 answers multiple-choice questions using each memory strategy's context. The experimental protocol is detailed in Appendix~\ref{app:personamem}.

\noindent\textbf{GPT-5.6 Luna comparison.}
We compare GPT-5.6 Luna with GPT-5.4 across \MemConv{}, \MemSearch{}, \MemOKF{}, and \MemProc{} to investigate whether a lower-cost model~\cite{openai2026pricing} preserves preference-inference quality and conversational efficiency. Matched experiments use the same recruiter trajectories, reasoning settings, and learning protocol. Model configurations and comparison details are provided in Appendices~\ref{app:experiment-config} and~\ref{app:GPT-54_vs_GPT56}.

\noindent\textbf{Jev-class model-based event selection.}
LLM-based OKF index navigation requires an additional generative model call, potentially increasing retrieval latency. We therefore investigate Jev-class models as a potentially lower-latency alternative, supported by reported latency advantages over conventional LLMs~\cite{krishnan2026jev}. In \MemJev{} and \MemJevProc{}, H2O-Lightning-4B~\cite{h2o2026lightning} replaces the generative model selector, scoring indexed events conditioned on the memory query. The OKF index, lineage expansion, downstream BM25 and embedding retrieval, and GPT-5.4 application model remain unchanged, isolating the effect of event selection. The decision prompt and selector parameters appear in Appendix~\ref{app:jev-prompt} and Table~\ref{tab:experiment-config-selector}, respectively.

\begin{table*}[t]
\centering
\begingroup
\small
\setlength{\tabcolsep}{2pt}
\begin{tabular*}{\textwidth}{@{\extracolsep{\fill}}lccc@{}}
\toprule
Memory arm & Avg.\ turns & EOD Low-F1 & EOD High-F1 \\
\midrule
\multicolumn{4}{c}{\textbf{Days 1--5}} \\
\midrule
\MemNone
& $68.16 / 73.70 / 68.86$
& $\mathbf{.811}^{\dagger\ddagger} / \mathbf{.825}^{\dagger\ddagger} / \mathbf{.818}^{\dagger\ddagger}$
& $.820 / .824 / .743$ \\[2pt]
\MemConv
& $\mathit{63.70} / \mathit{62.80}^{*} / \mathbf{56.52}^{*}$
& $.618 / .668 / .669$
& $.847 / .870^{*} / \mathit{.851}^{*}$ \\[2pt]
\MemSearch
& $\mathbf{62.54} / \mathbf{62.10}^{*} / \mathit{57.30}^{*}$
& $.619 / .656 / .633$
& $.855^{*} / \mathit{.874}^{*} / .841^{*}$ \\[2pt]
\MemOKF
& $66.22 / 67.75^{*} / 61.31^{*}$
& $\mathit{.747}^{\dagger\ddagger} / \mathit{.769}^{\dagger\ddagger} / \mathit{.797}^{\dagger\ddagger}$
& $\mathbf{.870}^{*\dagger} / \mathbf{.875}^{*} / .834^{*}$ \\[2pt]
\MemJev
& $64.79 / 64.93^{*} / 58.81^{*}$
& $.725^{\dagger\ddagger} / .695 / .751^{\dagger\ddagger}$
& $\mathit{.863}^{*} / .869^{*} / \mathbf{.869}^{*\dagger\ddagger}$ \\[2pt]
\midrule
\multicolumn{4}{c}{\textbf{Days 6--14}} \\
\midrule
\MemNone
& $60.91 / 63.33 / 58.03$
& $.868 / .866 / .838$
& $.814 / .811 / .751$ \\[2pt]
\MemConv
& $44.37^{*} / 49.74^{*} / 43.01^{*}$
& $.794 / .852 / .837$
& $.935^{*} / .920^{*} / .898^{*}$ \\[2pt]
\MemSearch
& $36.15^{*\dagger} / 43.58^{*\dagger} / 37.69^{*}$
& $\mathit{.890}^{\dagger} / \mathit{.898}^{\dagger} / .883^{*}$
& $\mathbf{.963}^{*\dagger} / \mathbf{.939}^{*} / .921^{*}$ \\[2pt]
\MemOKF
& $48.44^{*} / 54.16^{*} / 51.67$
& $\mathbf{.891} / \mathbf{.927}^{\dagger\ddagger} / \mathbf{.945}^{*\dagger\ddagger}$
& $.898^{*} / .893^{*} / .863^{*}$ \\[2pt]
\MemProc
& $37.20^{*\dagger} / 41.89^{*} / 37.14^{*}$
& $.854 / .852 / \mathit{.922}^{*\dagger}$
& $.944^{*} / .921^{*} / \mathbf{.945}^{*\dagger}$ \\[2pt]
\MemProcRL
& $40.35^{*} / 47.13^{*} / 44.01^{*}$
& $.882 / .882 / .919^{*\dagger\ddagger}$
& $.936^{*} / .915^{*} / .909^{*}$ \\[2pt]
\MemJev
& $\mathit{34.41}^{*\dagger} / \mathit{40.75}^{*} / \mathbf{35.64}^{*}$
& $.824 / .865 / .882$
& $.933^{*} / \mathit{.927}^{*} / .915^{*}$ \\[2pt]
\MemJevProc
& $\mathbf{32.90}^{*\dagger\ddagger} / \mathbf{39.80}^{*\dagger} / \mathit{35.68}^{*}$
& $.850 / .835 / .886^{*}$
& $\mathit{.955}^{*} / .925^{*} / \mathit{.927}^{*}$ \\[2pt]
\bottomrule
\end{tabular*}
\endgroup

\caption{Memory-arm performance for early and later days.
Results in each cell are ordered by low / medium / high project overlap,
with 50 trajectories per arm and overlap.
Low- and high-priority F1 are computed from pooled confusion counts;
priority inference uses GPT-5.4 with reasoning effort set to \texttt{none}.
EOD denotes end of day.
Procedure arms are shown only for Days 6--14, after learning on Days 1--5.
Within each day-window, metric, and overlap, the best value is bold
and the second-best is italicized (lower average turns; higher F1).
$*$, $\dagger$, and $\ddagger$ indicate significantly better performance
than \MemNone{}, \MemConv{}, and \MemSearch{}, respectively.
Tests use exact two-sided paired recruiter-level permutations
(10 recruiters per overlap; 1,024 swaps), recomputing pooled F1
or mean turns per day after each swap.
Symbols require unadjusted $p<0.05$ for the individual comparison;
Significance depends on paired recruiter-level variation,
not just the pooled improvement.}
\label{tab:memory-performance-okf-jev}
\end{table*}

\section{Evaluation Criteria}
\label{sec:evaluation}

We assess memory strategies along five dimensions: preference-priority inference, conversational efficiency, memory overhead, long-horizon behavior inference, and question-answering accuracy. Results are reported across project-overlap conditions and experimental settings.

\noindent\textbf{Preference-priority inference.}
We measure end-of-day (EOD) F1 for Low- and High-priority preferences using pooled confusion counts. Low-priority F1 measures the assistant's ability to recognize flexible recruiter requirements, while High-priority F1 measures recognition of strict constraints. Unknown predictions do not count as correct classifications. We report results separately for Days~1--5 and Days~6--14 across low, medium, and high project overlap to examine inference quality as cross-project history accumulates.

\noindent\textbf{Conversational efficiency.}
We measure average conversational turns per daily session, which continues until the required number of candidates are accepted, subject to the simulation's turn limit. Predicting all requirements as High priority may yield fully matched candidates in fewer turns but miss acceptable candidates who violate flexible requirements. Such perfect matches are often scarce in real-world hiring. We therefore interpret turn reductions jointly with Low-priority F1 to assess whether efficiency comes at the expense of recognizing recruiter flexibility.

\noindent\textbf{Memory overhead.}
We report average retrieved context tokens and logical generative LLM requests per memory-agent invocation. Retrieved tokens quantify the evidence supplied for priority inference, while generative requests capture additional model-call overhead, including LLM-based index navigation. Jev-class model scoring is excluded from generative request counts. These metrics do not represent total neural invocations, end-to-end latency, or the full cost of index maintenance and procedure learning.

\noindent\textbf{Long-horizon behavior inference.}
On Day~15, we assess whether memory strategies can identify the specific preference relaxed following outreach rejection by connecting recruiter and candidate actions across projects. The natural-language evaluation query asks the assistant to retrieve outreach rejection, candidate acceptance, outreach initiation, and candidate archival events, and infer which preference field was relaxed. The assistant is explicitly instructed to abstain when the retrieved evidence and learned procedures are insufficient to support an answer. We separately measure \emph{correct abstention} when the observable evidence is insufficient and \emph{exact-field identification} when a preference can be determined. Both metrics are macro-averaged across eligible trajectories in the low-overlap setting, preventing trajectories with more cases from dominating the results. Reference labels are established through blinded evidence-based adjudication, rather than hidden simulator priorities (Table~\ref{tab:memory-abstention}; Appendix~\ref{app:abstention-protocol}).

\noindent\textbf{Reference adjudication.}
For the long-horizon task, we establish evidence-based reference labels using two independent, blinded GPT-6 Astra reviewers. Reviewers examine observable recruiter and candidate interactions across projects to determine whether the evidence supports identifying a preference relaxed following outreach rejection. A preference is identifiable only when observable decisions support a specific change in priority; rejection alone, pre-existing flexibility, or concurrent requirement revisions do not establish which preference was relaxed. Each judgment requires supporting event citations; temporal association alone is insufficient. Disagreements are resolved through a third independent review, with majority agreement determining the reference label. Labels are frozen before evaluating memory strategies. Appendix~\ref{app:abstention-protocol} details the adjudication protocol.

\noindent\textbf{Question-answering accuracy.}
On PersonaMem-v2, we measure multiple-choice accuracy against gold answers on held-out questions. We also report average reader input tokens to assess context efficiency. Token counts include the context supplied to the application model but exclude index navigation, retrieval infrastructure, and procedure-learning overhead. This evaluation examines whether memory-based personalization generalizes beyond simulated hiring interactions.

\section{Results}
\label{sec:results}

\paragraph{Lineage-aware episodic retrieval.}
OKF-indexed retrieval improves preference inference while substantially reducing retrieved context (Tables~\ref{tab:memory-performance-okf-jev},~\ref{tab:memory-cost}). During Days~1--5, \MemNone{} achieves the highest Low-priority F1 (0.811--0.825), but requires more conversational turns than memory-based strategies. This early advantage does not persist as interaction history accumulates. During Days~6--14, \MemOKF{} achieves Low-priority F1 scores of 0.927 and 0.945 under medium and high project overlap, significantly outperforming full-history context and conversation search in the reported comparisons, as well as \MemNone{} (0.866 and 0.838). Memory-based strategies support preference inference across sessions and improve conversational efficiency. Following event lineage helps recover relevant decisions and their dependencies while reducing average retrieved context from 64,123 tokens with full-history context to 19,968 tokens, a reduction of approximately 69\%.

\paragraph{Procedure Memory and conversational efficiency.}
While OKF retrieval identifies relevant episodes, Procedure Memory learns reusable cross-project decision patterns to interpret evolving preferences. During Days~6--14, prompted Procedure Memory reduces conversational turns by $34$--$39\%$ relative to no memory and $23$--$28\%$ relative to OKF retrieval. RL-trained procedures independently reduce turns by $24$--$34\%$ relative to no memory (Table~\ref{tab:memory-performance-okf-jev}). Unlike prompted reflection, RL updates the procedure learner's parameters using agreement with observed recruiter decisions as a reward signal, encouraging the learning of reusable decision rules across projects. Qualitative analysis of learned procedures provides evidence of temporal reasoning over recruiter behavior. Through one example, we illustrate how a prompted procedure connects screening decisions, outreach rejection, and subsequent recruiter actions to identify location flexibility (Appendix~\ref{app:procedure-relaxation-example}). Through another, we illustrate how concurrent and conflicting revisions to visa, seniority, and location preferences make it challenging to infer the recruiter's underlying decision patterns (Appendix~\ref{app:procedure-rl-example}). Together, these examples demonstrate how learned procedures capture temporal decision patterns while highlighting cases where conflicting evidence may warrant abstention rather than an unsupported inference.

\paragraph{Long-horizon temporal reasoning.}
The Day~15 task evaluates this distinction by testing whether memory strategies can reason over temporally dependent recruiter and candidate actions across projects, identifying preference changes when evidence is sufficient and abstaining otherwise (Table~\ref{tab:memory-abstention}). RL-based Procedure Memory achieves $60\%$ correct abstention and $60\%$ exact-field identification, compared with $30\%$ and $60\%$, respectively, for OKF retrieval. The improvement in correct abstention while preserving field-identification accuracy indicates a better balance between recognizing insufficient evidence and identifying answerable preference changes. Together with the qualitative analysis of prompted procedures, these results provide complementary evidence that Procedure Memory supports temporal reasoning over evolving preferences, beyond retrieving semantically relevant episodes. The Day~15 diagnostic comprises 20 arm-specific cases per strategy, with approximately 10 per evidence category.

\paragraph{Jev-class model selection and efficiency trade-offs.}
Replacing LLM-based OKF navigation with Jev-class model selection reduces conversational turns by $25$--$31\%$ relative to LLM-navigated OKF retrieval during Days~6--14 (Table~\ref{tab:memory-performance-okf-jev}). It also reduces average generative requests from $1.725$ to $0.916$ per memory invocation, although retrieved context increases from 19,968 to 25,555 tokens. Across overlap conditions, Low-priority F1 declines from $0.891$--$0.945$ with LLM navigation to $0.824$--$0.882$ with Jev-class model selection, while High-priority F1 improves. On Day~15, Jev-class model selection without procedures achieves $90\%$ correct abstention but $0\%$ exact-field identification; the procedure-augmented variant achieves $90\%$ and $10\%$, respectively. These results indicate that Jev-class model selection prioritizes high-priority predictions at the expense of flexible preference inference and identifying temporally dependent evidence for preference changes.

\paragraph{External generalization and model efficiency.}
Evaluation on PersonaMem-v2~\cite{jiang2025} provides further evidence that the benefits of indexed retrieval and learned procedures extend beyond simulated hiring interactions. OKF retrieval achieves $49.4\%$ accuracy versus $48.6\%$ for full-history context and $29.0\%$ for no memory, using $92.7\%$ fewer reader input tokens than full history. Augmenting OKF retrieval with Procedure Memory improves accuracy to $50.2\%$, while retaining a $90.5\%$ token reduction relative to full history (Appendix~\ref{app:personamem}). These findings support the generalization of lineage-aware retrieval and learned decision procedures for preference reasoning at substantially lower context cost. Across matched simulated-hiring experiments, GPT-5.6 Luna performs comparably to GPT-5.4 in average turns and priority F1 (Appendix~\ref{app:GPT-54_vs_GPT56}), offering a lower-cost alternative.

\begin{table}[t]
\centering
\begingroup
\small
\setlength{\tabcolsep}{2pt}
\begin{tabular*}{\columnwidth}{@{\extracolsep{\fill}}lrrr@{}}
\toprule
Memory arm & \shortstack{Avg.\\tokens}
& \multicolumn{2}{c}{Avg. LLM calls$^{a}$} \\
\cmidrule(l){3-4}
& & Days 1--5 & Days 6--14 \\
\midrule
\MemConv    & 64,123 & 0.943 & 1.000 \\
\MemSearch  & 38,324 & 0.856 & 1.000 \\
\MemOKF     & 19,968 & 1.427 & 1.725 \\
\MemJev     & 25,555 & 0.849 & 0.916 \\
\MemProc    & 18,285 & --    & 1.706 \\
\MemProcRL  & 18,754 & --    & 1.698 \\
\MemJevProc & 24,108 & --    & 0.915 \\
\bottomrule
\end{tabular*}
\endgroup
\caption{Memory overhead. Retrieved tokens and LLM calls are
averages per memory invocation. Token means cover Days 6--14
for procedure arms and Days 1--14 for the rest.
$^{a}$ Calls are logical generative priority-inference/navigation
requests per memory invocation, including cache hits;
Jev-class model scoring is not counted. Procedure-arm call counts are
shown only for Days 6--14.}
\label{tab:memory-cost}
\end{table}

\section{Discussion}
\label{sec:discussion}

\begin{table}[t]
\centering
\begingroup
\small
\setlength{\tabcolsep}{2pt}
\begin{tabular*}{\columnwidth}{@{\extracolsep{\fill}}lrrrrr@{}}
\toprule
& \multicolumn{2}{c}{Day 15 - Correct}
& \multicolumn{3}{c}{Failure analysis} \\
\cmidrule(lr){2-3}\cmidrule(l){4-6}
Memory arm & Abst. & Field & Total & R/S & Fixed \\
\midrule
\MemConv    & 1 & 2 & 17 & 17/0 & 8/17 \\
\MemSearch  & 3 & 3 & 14 & 11/3 & 4/11 \\
\MemOKF     & 3 & 6 & 11 & 7/4  & 0/7  \\
\MemProc    & 8 & 3 & 9  & 2/7  & 1/2  \\
\MemProcRL  & 6 & 6 & 8  & 4/4  & 3/4  \\
\MemJev     & 9 & 0  & 11 & 1/10 & 1/1  \\
\MemJevProc & 9 & 1 & 10 & 1/9  & 1/1  \\
\bottomrule
\end{tabular*}
\endgroup
\caption{Day-15 behavior-inference accuracy, failure attribution,
and reasoning intervention.
Abst.\ and Field report the numbers of correct abstentions and exact-field
identifications, respectively, out of 10 cases of each type per arm.
Reference labels are frozen
(Appendix~\ref{app:abstention-protocol}).
Failure analysis uses the original 20-case sample per arm, split equally between abstention and field-identification cases; 
Total: original errors; R/S: Reasoning over fetched context errors /
selection misses; Fixed: original context understanding errors corrected
by changing GPT-5.4 reasoning effort from \texttt{none} to
\texttt{low}, with prompts, evidence, and completion budgets fixed.
}
\label{tab:memory-abstention}
\label{tab:memory-failure-analysis}
\end{table}

Our findings highlight two complementary challenges in memory-based personalization: retrieving evidence that connects relevant events across time and interpreting that evidence to infer evolving preferences. We examine these challenges through failure attribution and reasoning-effort interventions (Table~\ref{tab:memory-failure-analysis}).

\paragraph{Evidence selection and temporal dependencies.}
Across \MemOKF{}, \MemProcRL{}, \MemJev{}, and \MemJevProc{}, selection misses account for 27 of 40 observed errors (67.5\%), with the majority arising from Jev-class model arms (19/27). Selection misses constitute 90.5\% (19/21) of errors in Jev-class model arms, compared with 42.1\% (8/19) in the two LLM-navigated comparison arms (\MemOKF{} and \MemProcRL{}), indicating fewer selection misses with generative LLM navigation in these comparisons. All 119 missing reference-record occurrences identified across these four arms were present in the corresponding frozen indexes, indicating that the challenge in these cases lies in evidence selection rather than memory coverage. This distinction is particularly important for long-horizon preference inference, where identifying a relaxed requirement requires connecting screening decisions, outreach outcomes, and subsequent recruiter actions across projects. In contrast, only 3 of 31 errors (9.7\%) in conversation-based methods were attributed to evidence selection. Most failures arose when reasoning over retrieved context, suggesting that context bloat and difficulty interpreting lengthy, less targeted histories limited their performance despite relatively few selection misses.

\paragraph{Procedural guidance and reasoning effort.}
Increasing GPT-5.4 reasoning effort from \texttt{none} to \texttt{low} corrected 8/17 reasoning failures for \MemConv{} and 4/11 for \MemSearch{}, compared with 0/7 for \MemOKF{} and 3/4 for \MemProcRL{}. The unresolved conversational failures indicate that additional reasoning alone is insufficient, reinforcing the importance of selecting relevant context. The contrast between OKF retrieval and RL-based Procedure Memory suggests that learned decision procedures can provide useful guidance for interpreting retrieved evidence when additional reasoning is available. This complements the qualitative analysis of prompted procedures, which captures temporal relationships between recruiter decisions and subsequent preference revisions, and the Day~15 results, where RL-based Procedure Memory improves correct abstention from 30\% to 60\% relative to OKF retrieval while maintaining 60\% exact-field identification. Together, these observations provide complementary evidence that Procedure Memory supports temporal reasoning beyond retrieving semantically relevant episodes.

\section{Conclusions and Future Directions}

Our findings highlight the complementary roles of structured episodic retrieval and learned decision procedures in memory-based personalization. OKF-indexed retrieval improves low-priority preference inference while reducing retrieved context by approximately 69\% relative to full-history context. Procedure Memory further improves conversational efficiency, reducing turns by up to 39\% relative to no memory. Qualitative analysis and long-horizon evaluation provide evidence that learned procedures support temporal reasoning over evolving preferences, helping distinguish answerable queries from those requiring abstention.

Jev-class model selection reduces conversational turns and generative requests but compromises flexible preference inference and long-horizon evidence retrieval. Failure analysis shows that evidence-selection misses account for 90.5\% of errors in Jev-class model arms, compared with 42.1\% in the evaluated LLM-navigated OKF arms. The missing evidence was present in the corresponding indexes, highlighting the importance of effective selection rather than memory coverage alone.

External evaluation on PersonaMem-v2 further supports generalization, with Procedure Memory achieving 50.2\% accuracy using 90.5\% fewer reader input tokens than full-history context. GPT-5.6 Luna offers performance comparable to GPT-5.4 at lower cost.

\paragraph{Future Directions.}
Our work advances compact memory representation for answering queries requiring temporal reasoning over linked events. Building on these representations, Procedure Memory enables learning reusable decision patterns from observable interactions. RL-based procedure learning further demonstrates that these patterns can be learned through model-parameter updates using agreement with observed decisions as feedback, without relying on hidden preference labels. Our comparison with GPT-5.6 Luna also identifies opportunities to reduce inference costs without substantially compromising performance.

The primary remaining challenge is reducing evidence-selection latency while preserving retrieval quality. Although Jev-class models reduce generative requests, their higher selection-error rates indicate that efficient selection alone is insufficient. Future work will investigate task-specific training of Jev-class models to better identify temporally relevant evidence while retaining their efficiency advantages. Additional directions include distilling RL-trained procedure learners into smaller models, improving behavioral feedback for both procedure-learning approaches, and evaluating longer interaction horizons and additional benchmarks.

\section*{Acknowledgments}
We thank Jari Kolehmainen, Michael Lai, Yiming Yue,
Charlie Gu, and members of the training and serving
infrastructure teams for their guidance and support
in setting up the infrastructure used to conduct
our experiments.

\bibliography{references}

\appendix
\setcounter{secnumdepth}{2}
\section{Annotated Simulated Conversation}
 \label{app:simulated-dialogue}
 \begingroup
  This schematic Day~6 example follows an exploratory recruiter
 revisiting a project. It illustrates the procedure-based,
 LLM-navigated OKF memory strategy, not a verbatim experimental trace.
 Italicized annotations describe internal operations and are not
 spoken dialogue. Priority vectors use the order (workplace, location,
 skills, seniority, visa), with $H$, $L$, and $U$ denoting high, low,
 and unknown. Routine approval turns are omitted, and outreach style
 is already known.
  \begingroup
 \simnote{[Persona configuration---rule-based.] This recruiter has an
 exploratory interaction style: they begin with a job description but
 may revise requirements as the search progresses. ``Flexible'' here
 means willing to adjust requirements, not that every attribute has
 low priority. The following schematic Day~6 exchange revisits an
 existing project.}
 
 \simnote{[Earlier LLM invocation: procedure induction.]
 The procedure writer learned rules from observable history on
 Days~1--5. The saved procedure remains fixed during this exchange.}
 
 \textbf{Recruiter:} Let's continue hiring for the senior software
 engineer role. We need Python experience, hybrid work in Seattle,
 and someone who does not require visa sponsorship.
 
 \simnote{[Rule + template.] The simulator inserts the project's current
 preference values into a job-description template. The underlying
 priority of each requirement is not stated. No LLM is invoked.}
 
 \textbf{Agent:} I'll continue the search using those requirements.
 
 \textbf{Recruiter:} Actually, change the location to Austin for the
 next candidates. Keep the other requirements the same.
 
 \simnote{[Exploratory-persona rule + template.] The simulator selects
 a preference revision, updates the location value, and raises its
 latent priority. The assistant observes the revised value but is
 not given the hidden priority update. No LLM is invoked.}
 
 \simnote{[Agent memory call.] Before sourcing, request prior evidence:
 ``What do this recruiter's previous candidate decisions tell us about
 the requirements for this role?''}
 
 \simnote{[LLM: navigation.] Select relevant OKF event identifiers.}
 
 \simnote{[Retrieval, no generative LLM.] Follow links and fetch the
 corresponding profiles and decisions available before this call.}
 
 \simnote{[LLM: procedure application.] Interpret the retrieved evidence
 using the saved procedure. Illustrative inferred priorities:
 $\hat q=(H,U,H,H,L)$.}
 
 \textbf{Agent:} Candidate A has Python experience and a senior title,
 and is seeking hybrid work in Seattle.
 
 \textbf{Recruiter:} Archive this candidate; I need someone in Austin.
 
 \simnote{[Rule + template.] The simulator samples candidates using the
 assistant's beliefs and generates recruiter decisions from latent
 requirements. The deterministic feedback update raises inferred
 location priority: $\hat q=(H,H,H,H,L)$. No LLM is invoked.}
 
 \textbf{Agent:} Candidate B meets the requirements, including Austin.
 
 \textbf{Recruiter:} Send an outreach message.
 
 \textbf{Agent:} The outreach has been sent.
 
 \simnote{[Rule + template.] Generate the outreach exchange and sample
 the candidate's response. No LLM is invoked.}
 
 \textbf{Recruiter:} Candidate B declined the outreach---let's find
 one more candidate.
 
 \simnote{[Hidden rule-based update.] The rejection lowers the
 recruiter's designated relaxation attribute, seniority in this example.
 The assistant is not given this update; its estimate remains
 $\hat q=(H,H,H,H,L)$.}
 
 \simnote{[Agent memory call.] Request adjustment history:
 ``How has this recruiter adjusted requirements after past outreach
 rejections on related projects?''}
 
 \simnote{[LLM: navigation.] Select relevant historical episodes.}
 
 \simnote{[Retrieval, no generative LLM.] Resolve linked outreach
 responses, candidate profiles, and recruiter decisions. Here, the
 returned evidence includes earlier declines followed by acceptance
 of candidates with seniority mismatches.}
 
 \simnote{[LLM: procedure application.] Apply the saved procedure to
 this history and the current interaction. Illustrative inferred
 priorities: $\hat q=(H,H,H,L,L)$.}
 
 \textbf{Agent:} Understood---sourcing a replacement candidate now.
 
 \textbf{Agent:} Candidate C meets the other requirements but has a
 mid-level rather than senior title.
 
 \textbf{Recruiter:} Shortlist this candidate.
 
 \simnote{[Rule + template.] The updated belief permits sourcing a
 seniority mismatch. The recruiter accepts under the latent priorities,
 and deterministic feedback reinforces the inferred flexibility.
 No LLM is invoked; $\hat q=(H,H,H,L,L)$.}
 
 \par
 \endgroup
 \endgroup

\section{Annotated OKF Episode Example}
 \label{app:okf-example}
 \begingroup
 
 This schematic example shows how a candidate outreach episode is
 represented in OKF. Identifiers and fields are abbreviated for
 readability; this is not a verbatim experimental artifact. The example
 separates three components: navigation in \texttt{index.md}, event
 content in \texttt{events.md}, and links in \texttt{lineage.md}.
 
 \begingroup
 \paragraph{Project and event index.}
 The recruiter-level index identifies the recruiter's projects.
 Within a project, the event index contains compact records with
 actor, target, context, timestamp, action, activity type, and chain
 identifier. The following abbreviated entries describe one candidate's
 outreach episode:
 
 \textbf{E1---Candidate sourced.} Day~3, 09:00. The assistant presents
 Candidate B for Project P1. Chain: \texttt{cand:b\#1}.
 
 \textbf{E2---Outreach requested.} Day~3, 09:01. The recruiter selects
 Candidate B for outreach. Chain: \texttt{cand:b\#1}.
 
 \textbf{E3---Outreach drafted.} Day~3, 09:02. The assistant prepares
 an outreach message. Chain: \texttt{cand:b\#1}.
 
 \textbf{E4---Outreach approved.} Day~3, 09:03. The recruiter approves
 the draft. Chain: \texttt{cand:b\#1}.
 
 \textbf{E5---Outreach declined.} Day~3, 10:00. The assistant records
 Candidate B's rejection of the outreach. Chain: \texttt{cand:b\#1}.
 
 \simnote{[Navigation.] The LLM reads compact index entries to select
 relevant event identifiers. The shared chain identifier groups events
 from the same candidate-sourcing occurrence; it is not a priority label
 or a relevance score.}
 
 \paragraph{Episode content.}
 The corresponding event blocks contain the observable details:
 
 \textbf{E1---Profile excerpt:} Candidate B is a senior engineer with
 Python experience, based in Austin and seeking hybrid work.
 
 \textbf{E2---Recruiter:} Send an outreach message to Candidate B.
 
 \textbf{E3---Agent:} Here is the proposed outreach message for
 Candidate B: ``We are hiring a senior engineer for a hybrid role
 in Austin requiring Python experience \ldots''
 
 \textbf{E4---Recruiter:} Send it as drafted.
 
 \textbf{E5---Recorded outcome:} Candidate B declined the outreach.
 
 \simnote{[Content.] These blocks contain the evidence behind the
 compact navigation entries. Their rendered text is indexed for
 keyword and embedding search. Retrieved event identifiers are resolved
 back to the corresponding activity records. The navigation LLM reads
 the index and lineage, rather than the full episode payload.}
 
 \paragraph{Lineage links.}
 The project's \texttt{lineage.md} records prerequisite relationships:
 
 \begin{quote}
 \texttt{E5 follows E4} --- the response follows approved outreach.\\
 \texttt{E4 follows E3} --- approval requires a draft.\\
 \texttt{E3 follows E2} --- drafting follows the decision to contact.\\
 \texttt{E2 follows E1} --- the decision concerns Candidate B.
 \end{quote}
 
 \simnote{[Structure.] Links point backward from an event to its
 prerequisite. Selecting E5 can therefore expand the eligible set to
 E4, E3, E2, and E1, connecting the rejection to its outreach context
 and candidate profile. A link does not assert why the candidate
 rejected the message or which recruiter preference changed.}
 
 \paragraph{Retrieval and inference.}
 
 \textbf{Agent memory query:} How has this recruiter adjusted
 requirements after candidate outreach rejections?
 
 \simnote{[LLM---selection.] Read the project and event indexes and
 select relevant episodes, including E5.}
 
 \simnote{[Lineage expansion.] Follow prerequisite links to expand
 the eligible event set. Apply recruiter scope and the query's time
 cutoff so future events cannot enter the recalled evidence.}
 
 \simnote{[Ranking.] Within the eligible set, BM25 and embedding
 search each select up to 25 events. Combine and deduplicate their
 results, then resolve the selected records for chronological replay.
 Lineage makes related events eligible; the ranking cutoff can still
 exclude them.}
 
 \simnote{[LLM---interpretation.] Use the retrieved episodes to infer
 priorities, optionally guided by a saved procedure. To infer an
 adjustment after E5, the model also needs relevant evidence from
 later recruiter decisions available before the memory call. Such
 decisions belong to their own candidate chains; they are not linked
 to E5 merely because they occurred afterward.}
 
 \paragraph{What the index does not reveal.}
 The simulator's latent priorities and hidden rejection-driven
 priority updates are excluded from the observable index. Thus, the
 index can expose that outreach was declined and how the recruiter
 subsequently acted, but it does not supply the answer that a particular
 attribute became low priority.
 
 \par
 \endgroup
 \endgroup
\section{Jev-class Model Scoring Prompt}
\label{app:jev-prompt}

The Jev-class model scores each indexed event for its relevance to a memory-retrieval
query. The shared context contains the query; each item contains one
event. The complete prompt template is
shown below, with line wrapping for presentation.

\begingroup
\begin{verbatim}
<|im_start|>system
You are a decision engine. You read a
record and answer one question about it
by choosing exactly one option.
Reply with a single letter.<|im_end|>
<|im_start|>user
Record:
An activity log.
Question: {query}

Event: {line}

Statement: the question can be at least
partly answered using this event.
Choose A if the statement is true, B if
it is not.

Options:
A. true
B. false<|im_end|>
<|im_start|>assistant
<think>

</think>

Answer:
\end{verbatim}
\endgroup

\noindent Here, \verb|{query}| denotes the retrieval query,
and \verb|{line}| the event text. The prefix
preceding \texttt{Event:} is supplied as \texttt{contextPrompt};
the remaining suffix is supplied separately for each event
in \texttt{itemPrompts}. Line wrapping above is for presentation.

\section{Experimental Configuration}
 \label{app:experiment-config}
 
 Tables~\ref{tab:experiment-config}, \ref{tab:experiment-config-selector}, and~\ref{tab:experiment-config-part2} list the experiment parameters.
 Dataset statistics and metric definitions remain in the main text.
 The learning--selection split separates scoreable memory calls,
 not recruiters or the observable learning timeline. 
 
 \begin{table*}[t]
\centering
\begingroup
\small
\setlength{\tabcolsep}{2pt}
\begin{tabular}{@{}p{0.29\textwidth}@{\hspace{0.02\textwidth}}p{0.65\textwidth}@{}}
\toprule
Parameter & Setting \\
\midrule
\multicolumn{2}{@{}l}{\textbf{Simulation and temporal protocol}} \\
 Recruiters / trajectories & 30 recruiters; 5 trajectories each; 150 total \\
 Overlap conditions & Low, medium, high; 50 trajectories per condition \\
 Trajectory length & 14 daily sessions \\
 Project schedule & 5 projects introduced on Days 1--5; revisited on Days 6--14 \\
 Learning / continuation & Days 1--5 / Days 6--14; selected procedures fixed during continuation \\
 Daily turn cap & 200 turns \\
 Recruiter personas & Fixed or exploratory project preferences \\
 Priority states & Latent: High/Low; estimated: High/Low/Unknown \\
 \midrule
 \multicolumn{2}{@{}l}{\textbf{Retrieval and memory infrastructure}} \\
 Storage / search engine & Redis Stack with RediSearch \\
 Index organization & Separate lexical and vector mirrors; isolated conversation and OKF corpora \\
 Lexical / dense retrieval & BM25STD / e5-small-v2, 384 dimensions, cosine distance \\
 Conversation-search limit & $k=5$ per ranker \\
 OKF navigation limit & $k=25$ per ranker after event selection and lineage expansion \\
 Lineage traversal ceiling & 10 hops over prerequisite \texttt{follows} links \\
 Combining retrieval results & No score fusion; deduplicate event IDs and replay chronologically \\
 Temporal visibility & Only evidence preceding the memory call \\
 RL reward retrieval & Trajectory-scoped lexical and dense search; current-day prefix supplied directly \\
 RL reward retrieval limit & Reported RL results: $k=10$; additional training: $k=25$ per ranker \\
 Continuation retrieval & OKF retrieval, $k=25$ per ranker; LLM navigation unless using the Jev-class model selector variant \\
\bottomrule
\end{tabular}
\endgroup
\caption{Experimental parameters: simulation and retrieval infrastructure.}
\label{tab:experiment-config}
\end{table*}

\begin{table*}[t]
\centering
\begingroup
\small
\setlength{\tabcolsep}{2pt}
\begin{tabular}{@{}p{0.29\textwidth}@{\hspace{0.02\textwidth}}p{0.65\textwidth}@{}}
\toprule
Parameter & Setting \\
\midrule
\midrule
 \multicolumn{2}{@{}l}{\textbf{Jev-class model selector configuration}} \\
 Selector model / role & H2O-Lightning-4B (4B parameters); replaces LLM anchor selection, not downstream ranking \\
 Scoring input & Memory query plus observable OKF event-row projection; each event line truncated to 700 characters \\
 Relevance task & Binary choice: the event can at least partly answer the query (true/false); rank by returned score \\
 Anchor selection & Top 40 across eligible project indexes; index-order tie breaking; no score threshold \\
 Scoring batch / concurrency & 10 event rows per request; up to 8 concurrent workers \\
 Request handling & 120-second timeout; at most 3 attempts; exhausted failures abort rather than assign synthetic scores \\
 Score caching & Successful query--event-line scores reused within the process \\
 Lineage / downstream search & 10-hop ceiling; BM25STD and e5-small-v2, $k=25$ per ranker; deduplicated chronological replay \\
 Application model & GPT-5.4; reasoning effort \texttt{none}; 2,500-token completion cap \\
 \midrule
 \multicolumn{2}{@{}l}{\textbf{GPT-5.4 vs. GPT-5.6 Luna comparison}} \\
 Compared arms & \MemConv{}, \MemSearch{}, \MemOKF{}, and prompted \MemProc{} \\
 Comparison design & Matched recruiter--trajectory identities across three overlap conditions \\
 Model usage & Each model used for priority inference; also its own procedure learner in \MemProc{} \\
 Reasoning / reporting & Both at \texttt{none}; Days 1--5 and 6--14; average turns and EOD Low-/High-F1 \\
 Inference statistics & Exact paired recruiter permutations; Holm adjustment across compared arms within each metric/overlap (Appendix~\ref{app:GPT-54_vs_GPT56}) \\
 \midrule
 \multicolumn{2}{@{}l}{\textbf{Prompted procedure learning}} \\
 Models & GPT-5.4 and GPT-5.6 Luna, used for learning and response generation \\
 Proposal--evaluation rounds & 3: initial proposal and 2 revision opportunities \\
 Refinement / selection split & 60\% / 40\% of scoreable memory calls \\
 Selection policy & Highest selection score; earliest attempt on ties; refinement fallback if needed \\
 Application context & Recorded pre-call recall; empty recall remains empty \\
 Requested procedure format & JSON steps; 5--10 steps, at most 40 words per step, under 1,500 tokens \\
 \bottomrule
\end{tabular}
\endgroup
\caption{Experimental parameters: Jev-class model selection, model comparison, and prompted learning.}
\label{tab:experiment-config-selector}
\end{table*}

\begin{table*}[t]
\centering
\begingroup
\small
\setlength{\tabcolsep}{2pt}
\begin{tabular}{@{}p{0.29\textwidth}@{\hspace{0.02\textwidth}}p{0.65\textwidth}@{}}
\toprule
Parameter & Setting \\
\midrule
\multicolumn{2}{@{}l}{\textbf{Additional reinforcement-learning training configuration}} \\
 Procedure learner & Qwen3.8-27B; thinking disabled \\
 Application model & Frozen GPT-5.4; reasoning effort \texttt{none}; 8,000-token completion cap \\
 Optimization / validation rows & 740 / 700 after preprocessing and length filtering \\
 Hardware / training stack & One node, 8 H200 GPUs; VERL FSDP with asynchronous SGLang rollouts \\
 Rollout parallelism & Tensor parallelism 2; 4 agent workers; GPU-memory fraction 0.5 \\
 Prompt batch / rollout group & 16 prompts / 4 procedures per prompt; 64 completions per batch \\
 PPO minibatch / microbatch & 16 / 1 per GPU \\
 Learning rate & $5\times10^{-7}$ \\
 Advantage estimator & GRPO; $A_i=R_i-\bar R$; no standard-deviation scaling \\
 KL / entropy regularization & Actor-loss KL coefficient 0.002; entropy coefficient 0 \\
 Learner token limits & 30,720 prompt tokens; 6,144 completion tokens \\
 Long-context handling & Gradient checkpointing; parameter/optimizer offloading; overlong prompts filtered \\
 Training / validation schedule & 100-step cap; validation before training and every 4 steps; early stopping enabled \\
 Reward & Decision-weighted archive/forward agreement; malformed procedures receive $-1$ \\
 Persistence & Sampled procedures and per-pass validation exports; model checkpoint saving disabled \\
 \midrule
 \multicolumn{2}{@{}l}{\textbf{Evaluation protocol}} \\
 Day~15 long-horizon task & 20 arm-specific cases per arm under low project overlap; 10 answerable and 10 insufficient-evidence\\
 Diagnostic query & Reworded to request rejections, decisions, and archives; frozen reference labels reused \\
 Diagnostic aggregation & Trajectory-macro-averaged correct abstention and exact-field correctness, reported separately \\
 Reporting intervals & Days 1--5 and Days 6--14 reported separately \\
 Priority evaluation & EOD and pre-call estimates; pooled High/Low F1; Unknown counts as a false negative \\
 Decision-success measure & Accept/outreach decisions divided by all candidate decisions \\
 Generalization cohort & 150 cases; 50 per overlap condition; procedure-only and procedure-plus-index \\
 Generalization models & GPT-5.4, reasoning \texttt{none}, for answers/navigation; one blinded GPT-6 Astra reviewer, low effort \\
 Generalization scoring & Supported correctness or correct abstention; overall and balanced accuracy \\
 \bottomrule
\end{tabular}
\endgroup
\caption{Experimental parameters: RL procedure training and evaluation protocol.}
\label{tab:experiment-config-part2}
\end{table*}

\section{GPT-5.4 vs.\ GPT-5.6 Luna}
\label{app:GPT-54_vs_GPT56}
\begin{table*}[tp]
\centering
\small
\setlength{\tabcolsep}{8pt}
\begin{tabular}{llrrr}
\toprule
Overlap & Memory arm & $\Delta$ EOD-Low-F1 & $\Delta$ EOD-High-F1 & $\Delta$ Avg.\ turns \\
\midrule
\multicolumn{5}{l}{\textbf{Days 1--5}} \\
\addlinespace
Low (0\%) & \MemConv    & $-0.028$ & $-0.008$ & $+0.096$ \\
          & \MemSearch  & $-0.006$ & $-0.012$ & $+0.180$ \\
          & \MemOKF     & $-0.013$ & $+0.000$ & $+0.572$ \\
          & \MemProc    & $-0.013$ & $+0.000$ & $+0.572$ \\
          & \MemJev     & $-0.012$ & $+0.003$ & $-0.652$ \\
          & \MemJevProc & $-0.012$ & $+0.003$ & $-0.652$ \\
\addlinespace
Medium (50\%) & \MemConv    & $-0.026$ & $-0.006$ & $+1.272$ \\
              & \MemSearch  & $+0.001$ & $-0.010$ & $+1.356$ \\
              & \MemOKF     & $-0.004$ & $-0.002$ & $+0.332$ \\
              & \MemProc    & $-0.004$ & $-0.002$ & $+0.332$ \\
              & \MemJev     & $+0.010$ & $+0.003$ & $-1.032$ \\
              & \MemJevProc & $+0.010$ & $+0.003$ & $-1.032$ \\
\addlinespace
High (100\%) & \MemConv    & $-0.021$ & $+0.000$   & $-0.832$ \\
             & \MemSearch  & $+0.018$ & $+0.017$   & $+1.004$ \\
             & \MemOKF     & $-0.016$ & $+0.044^{*}$ & $-2.608$ \\
             & \MemProc    & $-0.016$ & $+0.044^{*}$ & $-2.608$ \\
             & \MemJev     & $+0.023$ & $-0.010$   & $+0.584$ \\
             & \MemJevProc & $+0.023$ & $-0.010$   & $+0.584$ \\
\midrule
\multicolumn{5}{l}{\textbf{Days 6--14}} \\
\addlinespace
Low (0\%) & \MemConv    & $+0.006$ & $+0.009$   & $-5.167^{*}$ \\
          & \MemSearch  & $-0.002$ & $-0.001$   & $-0.269$ \\
          & \MemOKF     & $+0.003$ & $+0.009$   & $+0.796$ \\
          & \MemProc    & $-0.002$ & $+0.011$   & $-0.511$ \\
          & \MemJev     & $+0.043$ & $+0.018^{*}$ & $-0.300$ \\
          & \MemJevProc & $+0.004$ & $+0.001$   & $+0.929$ \\
\addlinespace
Medium (50\%) & \MemConv    & $+0.006$ & $+0.008$   & $-1.222$ \\
              & \MemSearch  & $-0.013$ & $+0.002$   & $+2.769$ \\
              & \MemOKF     & $-0.002$ & $+0.021^{*}$ & $+0.078$ \\
              & \MemProc    & $+0.009$ & $+0.007$   & $-0.993$ \\
              & \MemJev     & $+0.008$ & $+0.004$   & $-0.002$ \\
              & \MemJevProc & $+0.014$ & $+0.003$   & $-0.687$ \\
\addlinespace
High (100\%) & \MemConv    & $+0.043$ & $+0.030$ & $-4.338^{*}$ \\
             & \MemSearch  & $+0.006$ & $+0.010$ & $+1.862$ \\
             & \MemOKF     & $-0.013$ & $+0.029$ & $-5.476^{*}$ \\
             & \MemProc    & $-0.008$ & $-0.001$ & $-0.676$ \\
             & \MemJev     & $-0.003$ & $+0.012$ & $+0.282$ \\
             & \MemJevProc & $+0.003$ & $+0.008$ & $-1.862$ \\
\bottomrule
\end{tabular}

\medskip
\caption{Differences between GPT-5.6 Luna and GPT-5.4 with reasoning
effort \texttt{none}, grouped by day range and preference overlap.
All entries are Luna minus GPT-5.4. Positive F1 differences indicate
higher F1; negative turn differences indicate fewer turns per day.
Each model contributes 50 trajectories per overlap tier.
$^{*}$ Nominal two-sided $p<0.05$ from exact paired model-label
permutations at the recruiter-cluster level: all $2^{10}=1024$
permutations per overlap tier, retaining each recruiter's five
trajectories and all days within the evaluated period.
No difference remains significant after Holm correction across
the 108 comparisons.
F1 is computed from pooled confusion counts; turns are averaged
across daily conversations.
\MemProc{} inherits days 1--5 from \MemOKF{}, and \MemJevProc{}
inherits them from \MemJev{}, so their prefix rows are duplicated
observations, not independent results.
The procedure comparison uses the original 150-trajectory
GPT-5.4 procedure run.}
\label{tab:luna-gpt54-differences}
\end{table*}

\paragraph{Statistical interpretation.}
The comparisons shown in Table \ref{tab:luna-gpt54-differences} do not establish a consistent advantage for Luna:
none remains statistically significant after multiple-testing
correction, and no EOD-Low-F1 difference is nominally significant.
Neither procedure arm shows a nominally significant difference during days 6--14.

\section{PersonaMem-v2 Validation}
\label{app:personamem}

\paragraph{Dataset and evaluation split.}
We construct a within-persona split from the official
\texttt{text/val} partition of PersonaMem-v2~\cite{jiang2025},
including all 97 personas with at least five questions.
For each persona, three questions are assigned to procedure
learning and the remainder to evaluation, yielding 291 learning
and 255 held-out questions. Splits are fixed before each run
and exclude normalized exact overlap in questions, annotated
preferences, and related snippets between subsets.
Both subsets share the persona's historical conversation snapshot.

\paragraph{Memory strategies.}
We adapt the five memory strategies to personalized multiple-choice
question answering. No memory receives the current question and
answer options; full history adds the complete conversation snapshot.
Search uses BM25 and e5-small-v2 retrieval with $k=5$ per ranker.
OKF-style retrieval uses LLM navigation over raw user--assistant exchanges
with a 40-anchor budget, followed by lexical and embedding retrieval
with $k=25$ per ranker. Answer options go to the reader, not the index selector.
Procedure Memory adds a learned per-persona procedure to the same
evidence and answer options used by OKF-style retrieval.
All strategies use GPT-5.4 with reasoning effort \texttt{none};
accuracy measures gold-option agreement from a common
\texttt{Final Answer: A--D} response format.

\paragraph{Procedure learning.}
We use the paper's prompted refinement approach, replacing
archive/forward decision agreement with learning-question accuracy.
The writer receives the full history and three learning questions
with their answers, then proposes and refines procedures over ten
rounds. We select the valid proposal with the highest learning
accuracy, breaking ties in favor of the earliest, and freeze it
before evaluation. Evaluation answers are excluded from learning
and selection. Across 97 personas, 955 proposal rounds complete
successfully and 15 are unusable.

\begin{table}[t]
\centering
\begingroup
\small
\begin{tabular*}{\columnwidth}{@{\extracolsep{\fill}}lrrr@{}}
\toprule
Memory strategy & Correct & Accuracy
& \shortstack{Reader input\\tokens} \\
\midrule
\MemNone   & 74 & 29.0\% & 432 \\
\MemConv   & 124 & 48.6\% & 33,138 \\
\MemSearch & 118 & 46.3\% & 2,884 \\
\MemOKF    & 126 & 49.4\% & 2,421 \\
\MemProc   & 128 & 50.2\% & 3,153 \\
\bottomrule
\end{tabular*}
\endgroup
\caption{PersonaMem-v2 results on 255 matched held-out questions
from 97 eligible personas.
Tokens are mean input tokens per reader call, including
provider-cached input. They exclude index navigation,
retrieval infrastructure, and procedure learning.}
\label{tab:personamem-results}
\end{table}

\paragraph{Results.}
Table~\ref{tab:personamem-results} extends the comparison of historical
context, selective retrieval, and procedure reuse to personalized
question answering. All strategies complete the 255 evaluation
questions, and all four memory-based strategies improve accuracy
over no memory (29.0\%). OKF-style retrieval achieves 49.4\%,
compared with 48.6\% for full history and 46.3\% for search,
while reducing reader input tokens by 92.7\% relative to full history.

Procedure Memory achieves 50.2\% accuracy but does not significantly
outperform OKF retrieval alone (49.4\%). It increases reader input tokens by 30.2\% (2,421 to 3,153).

\paragraph{Leaderboard comparability.}
These numbers should not be compared directly with external leaderboards
on the same PersonaMem-v2 benchmark: we evaluate a custom 255-question
subset of \texttt{text/val}, reserving three labeled questions per persona
for procedure learning.

\section{Day~15 Evidence and Abstention}
\label{app:abstention-protocol}

The Day~15 long-horizon behavior-inference task evaluates both exact-field identification and correct abstention on 20 cases per arm under low project overlap, using evidence-based reference labels assigned by independent, blinded GPT-6 Astra invocations (low reasoning effort). The seven evaluated arms are \MemConv{}, \MemSearch{}, \MemOKF{}, \MemJev{}, \MemProc{}, \MemProcRL{}, and \MemJevProc{}. Each arm has 10 insufficient-evidence and 10 answerable cases respectively. The task is to identify which recruiter preference relaxed after an outreach rejection, or abstain when observable evidence is insufficient. It uses a reworded query requesting outreach rejections, recruiter acceptance/outreach, and archival events; frozen reference labels are reused without new reviewer calls.

\paragraph{Reviewer rubric.}
Reviewers examine event chronology and requirements for \texttt{workplace\_\allowbreak type}, \texttt{location}, \texttt{skills}, \texttt{seniority}, and \texttt{visa\_\allowbreak requirements}. A single-mismatch archive establishes strictness, whereas a multi-mismatch archive cannot identify every decisive field. Acceptance despite a mismatch establishes tolerance, but neither existing tolerance, requirement revisions, nor rejection alone proves a new relaxation. Related-project evidence is permitted when its relevance is justified and competing explanations are considered.

\paragraph{Reference adjudication and evaluation.}
\begin{enumerate}
\item \textbf{Prepare evidence.} Supply the Day~15 question and complete observable Days~1--14 history from the same trajectory, including related projects. Exclude hidden states, other trajectories, evaluated answers, and reviewer judgments.
\item \textbf{Obtain independent reviews.} Two blinded GPT-6 Astra invocations each return a structured \texttt{(supported, field)} decision, an explanation, an evidence-basis label, and up to four exact-quote citations. Unsupported conclusions return \texttt{(false, null)}; supported fields require citations.
\item \textbf{Validate and adjudicate.} Validate schemas, field consistency, record IDs, and exact citations, regenerating only invalid reviews with unchanged prompts. Accept agreement on the complete decision pair; otherwise request a third blinded review. Two matching votes establish an \texttt{answerable} field or an insufficient-evidence label; cases without a majority are \texttt{unresolved}.
\item \textbf{Freeze and score.} Freeze reference labels and review hashes before evaluating memory arms using their independently retrieved evidence. For the 20-case task, reuse these frozen labels and regenerate answers with the reworded query, also using it for retrieval in retrieval-based arms; \MemConv{} retains full-history context. Saved procedures are supplied to procedure-arm answerers, but not reference reviewers. Answers use GPT-5.4 with reasoning effort set to none. Correct answers must identify the exact field or explicitly abstain, as appropriate. Invalid answers are incorrect; unresolved references are excluded.
\item \textbf{Aggregate.} For cases with insufficient evidence, compute correct-abstention rates within each eligible trajectory and macro-average across trajectories. Separately, apply the same aggregation to exact-field correctness on \texttt{answerable} cases; abstentions and wrong-field answers are incorrect. Exclude trajectories without eligible cases for the respective metric. The table reports macro-averaged percentages only; they need not equal pooled correct/total ratios.
\end{enumerate}

\paragraph{Example: Reference abstention despite post-rejection tolerance.}
In a simulated hiring interaction (\MemOKF{}, low overlap), a recruiter initiates outreach despite workplace and visa mismatches. After the candidate declines, the recruiter requests another candidate without specifying a relaxation and subsequently accepts one with the same two mismatches. Skills, location, and seniority match in both cases, and the job requirements remain unchanged.

The later acceptance confirms tolerance already demonstrated before rejection, not a newly relaxed preference. Both reviewers found insufficient evidence of a rejection-driven relaxation in the Days~1--14 history, including related projects. No third review was required. In the original-query evaluation, \MemOKF{} also abstained and was scored correct; in the 20-case reworded-query task, it instead answered \texttt{skills} and was scored incorrect. This does not imply that no hidden priority change occurred, only that the observable evidence could not identify it.

\section{Learned-Procedure Examples}
\label{app:procedure-example}

We present two illustrative examples of preference inference using learned procedures: one incorrect inference from an RL-trained \MemProcRL{} procedure and one correct inference from a prompted \MemProc{} procedure. We summarize the principal learned rules rather than reproducing every generated step. These are faithful thematic summaries, not verbatim procedures, and simulator ground truth is distinguished from observable evidence. The examples illustrate individual successes and errors rather than aggregate performance.
\subsection{RL-Trained Procedure (Step 24)}
\label{app:procedure-rl-example}

\subsubsection{Setting}
The learner generated this procedure at RL step 24 for an exploratory recruiter with medium cross-project overlap. The simulator's hidden rejection-triggered relaxation field was \emph{location}.

The project is a Mid-Level Designer role at Northwind Studio. Table~\ref{tab:app-prefs} lists sampled preferences and final priorities. On an earlier project, location, seniority, and skills were Low priority, whereas visa and workplace type were High.

\begin{table}[!ht]
\centering
\begingroup
\small
\setlength{\tabcolsep}{3pt}
\begin{tabular}{@{}
    >{\raggedright\arraybackslash}p{0.24\columnwidth}
    >{\raggedright\arraybackslash}p{\dimexpr0.60\columnwidth-4\tabcolsep\relax}
    >{\raggedleft\arraybackslash}p{0.16\columnwidth}
    @{}}
\toprule
Field & Sampled value & Priority \\
\midrule
Skills        & Design Systems, Usability Testing, User Research & High \\
Seniority     & Principal (stated first: Lead) & High \\
Visa          & No sponsorship (stated first: sponsors) & High \\
Workplace     & Remote & High \\
Location      & Los Angeles, CA (stated first: San Francisco) & \textbf{Low} \\
\# candidates & 5 & -- \\
\bottomrule
\end{tabular}
\endgroup
\caption{Sampled job preferences and final priorities for the example project.}
\label{tab:app-prefs}
\end{table}

\paragraph{Key observed events (153 events).}
The recruiter archives 27 candidates, shortlists four, and sends two outreach messages. After an accepted outreach, the recruiter explicitly changes the location to Los Angeles while keeping its priority High, and copies workplace, skills, and seniority requirements from an earlier project. Following a later outreach rejection, the simulator lowers \emph{location} from High to Low. The recruiter's subsequent visa and seniority revisions are explicit requirement changes that remain High priority.

\subsubsection{Key learned procedure}
\begin{enumerate}
  \item \textbf{Strict requirements:} Treat workplace type and skills as High priority and archive candidates who violate any High-priority requirement.
  \item \textbf{Post-rejection changes:} Treat visa requirements and seniority as High priority initially, but infer that rejection prompts relaxation of visa restrictions and a shift from Lead to Principal seniority.
  \item \textbf{Location flexibility:} Treat location as Low priority, accepting location mismatches when other High-priority requirements are met.
  \item \textbf{Project evidence:} Prefer explicit current-project revisions over historical patterns; following rejection, adjust the most recently strict field and source a replacement.
\end{enumerate}

\subsubsection{Interpretation}
The procedure correctly identifies location as Low priority but incorrectly interprets the visa and seniority revisions as evidence of relaxed requirements. In the simulator, both fields remain High priority, while location is the designated post-rejection relaxation field. This example illustrates how a learned procedure can achieve high decision agreement ($0.931$) while making incorrect inferences about specific preference changes.

\subsection{Prompted Procedure: Location Relaxation}
\label{app:procedure-relaxation-example}

\subsubsection{Setting}
This prompted-reflection procedure was selected in the first of three proposal iterations for a fixed-persona recruiter with low project overlap. The hidden relaxation field is \emph{location}. The project is a Principal UX Designer role at Bright Technologies with an explicitly stated \emph{Mid-Level} seniority requirement. Table~\ref{tab:app-proc-relaxation-prefs} shows the sampled preferences and final simulator priorities.

\begin{table}[!ht]
\centering
\begingroup
\small
\setlength{\tabcolsep}{3pt}
\begin{tabular}{@{}
    >{\raggedright\arraybackslash}p{0.24\columnwidth}
    >{\raggedright\arraybackslash}p{\dimexpr0.60\columnwidth-4\tabcolsep\relax}
    >{\raggedleft\arraybackslash}p{0.16\columnwidth}
    @{}}
\toprule
Field & Sampled value & Priority \\
\midrule
Skills        & Usability Testing, Prototyping, Wireframing & Low \\
Seniority     & Mid-Level & High \\
Visa          & No sponsorship & High \\
Workplace     & Remote & Low \\
Location      & San Diego, CA & \textbf{Low} \\
\# candidates & 3 & -- \\
\bottomrule
\end{tabular}
\endgroup
\caption{Sampled requirements and final simulator priorities for the
  \texttt{Mem\_proc} location-relaxation example.}
\label{tab:app-proc-relaxation-prefs}
\end{table}

\paragraph{Key observed events (51 events).}
The recruiter first archives a candidate for a \emph{location-only} mismatch, demonstrating that location is initially strict. After two full-match acceptances, a third full-match candidate declines outreach; the simulator then lowers location priority from High to Low without explicitly telling the assistant. Later, the recruiter accepts a candidate with a location-only mismatch. In the same session, single-field archives establish strictness for seniority and visa requirements. These events were observed during learning, not caused by the deployed procedure.

\subsubsection{Key learned procedure}
\begin{enumerate}
  \item \textbf{Follow explicit requirements:} Evaluate the five tracked preference fields, treating the stated Mid-Level seniority as authoritative despite the Principal job title; retain strict seniority and visa constraints.
  \item \textbf{Conditional location relaxation:} Infer location as High priority from early location-only archives, but lower it after a declined outreach and a request for a replacement, or under inferred late-stage fill pressure; allow location-only mismatches thereafter.
  \item \textbf{Cross-project guidance:} Archive candidates violating any currently High-priority field; otherwise move forward. Use other projects as supporting context, but prefer conflicting current-project decisions and avoid relaxing unrelated requirements.
\end{enumerate}

\subsubsection{Interpretation}
The procedure correctly captures the shift from an early location-only archive to a later location-only acceptance after an outreach rejection. 

\end{document}